\documentclass[utf8,final]{frontiersSCNS}

\usepackage{url,hyperref,lineno,microtype,subcaption}
\usepackage[onehalfspacing]{setspace}
\usepackage{tikz-qtree}
\usepackage{multirow}

\def\cc{\otimes}
\def\ccorr{\oplus}

\def\R{\mathbb R}

\def\vec#1{\mathbf{#1}}

\def\cos#1#2{cos(#1,#2)}

\def\matrix#1{\mathbf{#1}}

\DeclareMathOperator*{\argmax}{arg\,max}

\usepackage{kbordermatrix}

\def\keyFont{\fontsize{8}{11}\helveticabold }
\def\firstAuthorLast{Ferrone\&Zanzotto} 
\def\Authors{Lorenzo Ferrone\,$^{1}$ and  Fabio Massimo Zanzotto\,$^{1,*}$}
\def\Address{$^{1}$Department of Enterprise Engineering, University of Rome Tor Vergata, Rome, Italy}
\def\corrAuthor{Fabio Massimo Zanzotto}

\def\corrEmail{fabio.massimo.zanzotto@uniroma2.it}

\begin{document}
\onecolumn
\firstpage{1}

\title[Running Title]{Symbolic, Distributed and Distributional Representations for Natural Language Processing in the Era of Deep Learning: a Survey} 

\author[\firstAuthorLast ]{\Authors} 
\address{} 
\correspondance{} 

\extraAuth{}

\maketitle

\begin{abstract}

\section{}
Natural language is inherently a discrete symbolic representation of human knowledge.  Recent advances in machine learning (ML) and in natural language processing (NLP) seem to contradict the above intuition: discrete symbols are fading away, erased by vectors or tensors called \emph{distributed} and \emph{distributional representations}. 
However, there is a strict link between distributed/distributional representations and discrete symbols, being the first an approximation of the second.  
A clearer understanding of the strict link between distributed/distributional representations and symbols may certainly lead to radically new deep learning networks. 
In this paper we make a survey that aims to renew the link between symbolic representations and distributed/distributional representations. This is the right time to revitalize the area of interpreting how discrete symbols are represented inside neural networks.

\tiny
 \keyFont{ \section{Keywords:} keyword, keyword, keyword, keyword, keyword, keyword, keyword, keyword} 
\end{abstract}

\section{Introduction}

Natural language is inherently a discrete symbolic representation of human knowledge. Sounds are transformed in letters or ideograms and these discrete symbols are composed to obtain words. Words then form sentences and sentences form texts, discourses, dialogs, which ultimately convey knowledge, emotions, and so on. This composition of symbols in words and of words in sentences follow rules that both the hearer and the speaker know \citep{Chomsky1957}. Hence, thinking to natural language understanding systems, which are not based on discrete symbols, seems to be extremely odd.

Recent advances in machine learning (ML) applied to natural language processing (NLP) seem to contradict the above intuition: discrete symbols are fading away, erased by vectors or tensors called \emph{distributed} and \emph{distributional representations}. 
In ML applied to NLP, \emph{distributed representations} are pushing deep learning models \citep{lecun2015deep,schmidhuber2015deep} towards amazing results in many high-level tasks such as  
image generation \citep{goodfellow2014generative}, image captioning \citep{vinyals2015show,xu2015show}, machine translation \citep{bahdanau2014neural,zou2013bilingual}, syntactic parsing \citep{NIPS2015_5635,weiss2015structured} and in a variety of other NLP tasks \cite{BERT}.
In a more traditional NLP, \emph{distributional representations} are pursued as a more flexible way to represent semantics of natural language, the so-called \emph{distributional semantics} (see \citep{DBLP:journals/jair/TurneyP10}). Words as well as sentences are represented as vectors or tensors of real numbers
Vectors for words are obtained observing how rhese words co-occur with other words in document collections.
Moreover, as in traditional compositional representations, vectors for phrases \citep{mitchell-lapata:2008:ACLMain,baroni-zamparelli:2010:EMNLP,ClarkCoeckeSadrzadeh2008,Grefenstette:2011:ESC:2145432.2145580,fabio2010-2} and sentences \citep{SocherEtAl2011:PoolRAE,SocherEtAl2012:MVRNN,kalchbrenner2013recurrent} are obtained by composing vectors for words. 

The success of distributed and distributional representations over symbolic approaches is mainly due to the advent of new parallel paradigms that pushed neural networks \citep{rosenblatt1958,werbos1974beyond} towards deep learning \citep{lecun2015deep,schmidhuber2015deep}.  Massively parallel algorithms running on Graphic Processing Units (GPUs) \citep{chetlur2014cudnn,cui2015scalable} crunch vectors, matrices and tensors faster than decades ago. 
The back-propagation algorithm can be now computed for complex and large neural networks. Symbols are not needed any more during ``resoning'', that is, the neural network learning and its application. Hence, discrete symbols only survive as inputs and outputs of these wonderful learning machines.

However, there is a strict link between distributed/distributional representations and symbols, being the first an approximation of the second \citep{FODOR19883,Plate:1994,Plate1995,DBLP:conf/slsp/FerroneZC15}.  The representation of the input and the output of these networks is not that far from their internal representation. The similarity and the interpretation of the internal representation is clearer in image processing \citep{10.1007/978-3-319-10590-1_53}. In fact, networks are generally interpreted visualizing how subparts represent salient subparts of target images. Both input images and subparts are tensors of real number. Hence, these networks can be examined and understood.
The same does not apply to natural language processing with its discrete symbols. 

A clearer understanding of the strict link between distributed/distributional representations and discrete symbols is needed \citep{Jang2018,Jacovi2018} to understand how neural networks treat information and to propose novel deep learning architectures. Model interpretability is becoming an important topic in machine learning in general \citep{Lipton2016}. This clearer understanding is then the dawn of a new range of possibilities: understanding what part of the current symbolic techniques for natural language processing have a sufficient representation in deep neural networks; and, ultimately, understanding  whether a more brain-like model -- the neural networks -- is compatible with methods for syntactic parsing or semantic processing that have been defined in these decades of studies in computational linguistics and natural language processing. There is thus a tremendous opportunity to understand whether and how symbolic representations are used and emitted in a brain model.

In this paper we make a survey that aims to draw the link between symbolic representations and distributed/distributional representations. This is the right time to revitalize the area of interpreting how symbols are represented inside neural networks. In our opinion, this survey will help to devise new deep neural networks that can exploit existing and novel symbolic models of classical natural language processing tasks.

The paper is structured as follow: first we give an introduction to the very general concept of representations and the difference between \emph{local} and \emph{distributed} representations \citep{Plate1995}.
After that we present each techniques in detail.
Afterwards, we focus on distributional representations \citep{DBLP:journals/jair/TurneyP10}, which we treat as a specific example of a distributed representation.
Finally we discuss more in depth the general issue of compositionality, analyzing three different approaches to the problem: compositional distributional semantics \citep{ClarkCoeckeSadrzadeh2008,baroni2014frege}, holographic reduced representations \citep{Plate:1994,neumann2001holistic}, and recurrent neural networks \citep{kalchbrenner2013recurrent,SocherEtAl2012:MVRNN}.

\section{Symbolic and Distributed Representations: Interpretability and \emph{Concatenative} Compositionality}

\emph{Distributed representations} put symbolic expressions in metric spaces where similarity among examples is used to learn regularities for specific tasks by using neural networks or other machine learning models. 
Given two symbolic expressions, their distributed representation should capture their similarity along specific features useful for the final task. For example, two sentences such as $s_1$=\emph{``a mouse eats some cheese''} and  $s_2$=\emph{``a cat swallows a mouse''} can be considered similar in many different ways: (1) number of words in common; (2) realization of the pattern ``\texttt{ANIMAL EATS FOOD}''. The key point is to decide or to let an algorithm decide which is the best representation for a specific task.

\emph{Distributed representations} are then replacing long-lasting, successful \emph{discrete symbolic representations} in representing knowledge for learning machines but these representations are less human \emph{interpretable}. Hence, discussing about basic, obvious properties of \emph{discrete symbolic representations} is not useless as these properties may guarantee success to distributed representations similar to the one of discrete symbolic representations. 


Discrete symbolic representations are human \emph{interpretable} as \emph{symbols are not altered in expressions}. This is one of the most important, obvious feature of these representations. Infinite sets of expressions, which are sequences of symbols, can be \emph{interpreted} as these expressions are obtained by concatenating a finite set of basic symbols according to some concatenative rules. During concatenation, symbols are not altered and, then, can be recognized. By using the principle of \emph{semantic compositionality}, 
the meaning of expressions can be obtained by combining the meaning of the parts and, hence, recursively, by combining the meaning of the finite set of basic symbols. 
For example, given the set of basic symbols $\mathcal{D}$ = \{\emph{mouse},\emph{cat},\emph{a},\emph{swallows},\emph{(},\emph{)}\}, expressions like $s_1$=\emph{``a cat swallows a mouse''} or $t_1$=\emph{((a cat) (swallows (a mouse)))} are totally plausible and interpretable given rules for producing natural language utterances or for producing tree structured representations in parenthetical form, respectively. This strongly depends on the fact that individual symbols can be recognized.

Distributed representations instead seem to \emph{alter symbols} when applied to symbolic inputs and, thus, are less interpretable. In fact, symbols as well as expressions are represented as vectors in these metric spaces. Observing distributed representations, symbols and expressions do not immediately emerge. Moreover, these distributed representations may be transformed by using matrix multiplication or by using non-linear functions. Hence, it is generally unclear: (1) what is the relation between the initial symbols or expressions and their distributed representations and (2) how these expressions are manipulated during matrix multiplication or when applying non-linear functions. In other words, it is unclear whether symbols can be recognized in distributed representations.

Hence, a debated question is whether discrete symbolic representations and distributed representations are two very different ways of encoding knowledge because of the difference in \emph{alterning symbols}. The debate dates back in the late 80s. For \citet{FODOR19883}, distributed representations in Neural Network architectures are \emph{``only an implementation of the Classical approach''} where classical approach is related to discrete symbolic representations. Whereas, for \citet{Chalmers1992}, distributed representations give the important opportunity to reason \emph{``holistically''} about encoded knowledge. This means that decisions over some specific part of the stored knowledge can be taken without retrieving the specific part but acting on the whole representation. However, this does not solve the debated question as it is still unclear what is in a distributed representation.  

To contribute to the above debated question, \citet{Gelder1990} has formalized the property of \emph{altering symbols in expressions} by defining two different notions of compositionality: \emph{concatentative} compositionality and \emph{functional} compositionality.  \emph{Concatenative compositionality} explains how discrete symbolic representations compose symbols to obtain expressions. In fact, the mode of combination is an extended concept of juxtaposition that provides a way of linking successive symbols without altering them as these form expressions. Concatenative compositionality explains discrete symbolic representations no matter the means is used to store expressions: a piece of paper or a computer memory.  Concatenation is sometime expressed with an operator like $\circ$, which can be used in a infix or prefix notation, that is a sort of function with arguments $\circ(w_1, ..., w_n)$. By using the operator for concatenation, the two above examples $s_1$ and $t_1$ can be represented as the following:
$$
a \circ cat \circ swallows \circ a \circ  mouse
$$
that represents a sequence with the infix notation and 
$$
\circ(\circ(a,cat),\circ(swallows,\circ(a,mouse)))
$$
that represents a tree with the prefix notation.
\emph{Functional compositionality} explains distributed representations. In functional compositionality, the mode of combination is a function $\Phi$ that gives a reliable, general process for producing expressions given its constituents. Within this perspective, semantic compositionality 
is a special case of functional compositionality where the target of the composition is a way to represent meaning \citep{newHypComp2003}. 

\emph{Local distributed representations} (as referred in \citep{Plate1995}) or \emph{one-hot encodings} are the easiest way to visualize how \emph{functional compositionality} act on \emph{distributed representations}. Local distributed representations give a first, simple encoding of discrete symbolic representations in a metric space. Given a set of symbols $\mathcal{D}$, a local distributed epresentation maps the $i$-th symbol in $\mathcal{D}$ to the $i$-th base unit vector $\vec{e}_i$ in $\R^n$, where $n$ is the cardinality of $\mathcal{D}$. Hence, the $i$-th unit vector represents the $i$-th symbol. 
In \emph{functional compositionality}, expressions $s = w_1\ldots w_k$ are represented by vectors $\vec{s}$ obtained with an eventually recursive function $\Phi$ applied to vectors $\vec{e}_{w_1} \ldots \vec{e}_{w_k}$. The function $f$ may be very simple as the sum or more complex. In case the function $\Phi$ is the sum, that is:
\begin{equation}
\vec{func}_{\Sigma}(s)= \sum_{j=1}^k \vec{e}_{w_j}
\label{func_sum}
\end{equation}
the derived vector is the classical bag-of-word vector space model \citep{Salton:1989}.  Whereas, more complex functions $f$ can range from different vector-to-vector operations like circular convolution in Holographic Reduced Representations \citep{Plate1995} to matrix multiplications plus non linear operations in models such as in recurrent neural networks \citep{Schuster:1997:BRN:2198065.2205129,hochreiter1997long} or in neural networks with attention \citep{AttentionIsAllYouNeed,BERT}. 
The above example can be useful to describe \emph{concatenative} and \emph{functional} compositionality. The set $\mathcal{D}$= \{\emph{mouse},\emph{cat},\emph{a},\emph{swallows},\emph{eats},\emph{some},\emph{cheese},\emph{(},\emph{)}\} may be represented with the base vectors $\vec{e}_i \in \R^9$ where $\vec{e}_1$ is the base vector for \emph{mouse}, $\vec{e}_2$ for \emph{cat},  $\vec{e}_3$ for \emph{a}, $\vec{e}_4$ for \emph{swallaws}, $\vec{e}_5$ for \emph{eats}, $\vec{e}_6$ for \emph{some}, $\vec{e}_7$ for \emph{cheese}, $\vec{e}_8$ for \emph{(}, and $\vec{e}_9$ for \emph{)}.
The additive functional composition of the expression $s_1$=\emph{a cat swallows a mouse} is then:
\begin{center}
\begin{tiny}
\begin{tabular}{|c|c|}
\emph{\textbf{expression in $e_i$}}  & \emph{\textbf{additive functional composition}} \\
\begin{tabular}{ccccc}
a & cat & swallows & a & mouse \\
$\begin{pmatrix}0 \\ 0 \\1 \\0 \\0 \\0 \\0 \\0 \\0\\\end{pmatrix}$ & $\begin{pmatrix}0\\1\\0\\0\\0\\0\\0 \\0 \\0\\\end{pmatrix}$ & $\begin{pmatrix}0\\0\\0\\1\\0\\0\\0 \\0 \\0\\\end{pmatrix}$ & $\begin{pmatrix}0 \\ 0 \\1 \\0 \\0 \\0 \\0 \\0 \\0\\\end{pmatrix}$ & $\begin{pmatrix}1\\ 0 \\0 \\0 \\0 \\0\\0 \\0 \\0 \\\end{pmatrix}$
\end{tabular}
&
\begin{tabular}{c}
$\vec{e_3} + \vec{e_2} + \vec{e_4} + \vec{e_3} + \vec{e_1}$
\\
$
\vec{func_{\Sigma}(s_1)}= \begin{pmatrix}   1\\1\\2 \\1\\0\\0 \\0\\0 \\0\end{pmatrix}
$
\end{tabular}
\end{tabular}
\end{tiny}
\end{center}
where the concatenative operator $\circ$ has been substituted with the sum $+$. Just to observe, in the additive functional composition $\vec{func_{\Sigma}(s_1)}$, symbols are still visible but the sequence is lost. Hence, it is difficult to reproduce the initial discrete symbolic expression. However, for example, the additive composition function gives the possibility to compare two expressions. Given the expression $s_1$ and $s_2$=\emph{a mouse eats some cheese}, the dot product between $\vec{func_{\Sigma}(s_1)}$ and $\vec{func_{\Sigma}(s_2)}=\begin{pmatrix}   1&0&1&0&1&1 &1&0 &0\end{pmatrix}^T$ counts the common words between the two expressions.  In a functional composition with a function $\Phi$, the expression $s_1$ may become $\vec{func_{\Phi}(s_1)} = \Phi(\Phi(\Phi(\Phi(\vec{e_3},\vec{e_2}),\vec{e_4}),\vec{e_3}),\vec{e_1})$ by following the concatenative compositionality of the discrete symbolic expression. The same functional compositional principle can be applied to discrete symbolic trees as $t_1$ by producing this distributed representation $\Phi(\Phi(\vec{e_3},\vec{e_2}),\Phi(\vec{e_4},\Phi(\vec{e_3},\vec{e_1})))$. Finally, in the functional composition with a generic recursive function $\vec{func_{\Phi}(s_1)}$, the function $\Phi$ will be crucial to determine whether symbols can be recognized and sequence is preserved.

\emph{Distributed representations} in their general form are more ambitious than distributed \emph{local} representations and tend to encode basic symbols of $\mathcal{D}$ in vectors in $\R^d$ where $d<<n$. These vectors generally alter symbols as there is not a direct link between symbols and dimensions of the space. Given a distributed local representation $\vec{e}_w$ of a symbol $w$, the encoder for a distributed representation is a matrix $\matrix{W_{d \times n}}$ that transforms  $\vec{x}_w$  in $\vec{y}_w = \matrix{W_{d \times n}} \vec{e}_w$. As an example, the encoding matrix $\matrix{W_{d \times n}}$ can be build by modeling words in $\mathcal{D}$ around three dimensions: number of vowels, number of consonants and, finally, number of non-alphabetic symbols.
Given these dimensions, the matrix $\matrix{W_{3 \times 9}}$ for the example is :
\begin{displaymath}
\matrix{W_{3 \times 9}}=
\begin{pmatrix}
3 &  1 & 1 & 2 & 2 & 2 & 3 & 0 & 0\\
2 &  2 & 0 & 6 & 2 & 2 & 3 & 0 & 0\\
0 &  0 & 0 & 0 & 0 & 0 & 0 & 1 & 1\\
\end{pmatrix}
\end{displaymath}
This is a simple example of a \emph{distributed} representation.  In a distributed representation \citep{Plate1995,Hinton:1986} the informational content is distributed (hence the name) among multiple units, and at the same time each unit can contribute to the representation of multiple elements. Distributed representation has two evident advantages with respect to a distributed local representation: it is more efficient (in the example, the representation uses only 3 numbers instead of 9) and it does not treat each element as being equally different to any other. In fact, \emph{mouse} and \emph{cat} in this representation are more similar than \emph{mouse} and \emph{a}. In other words, this representation captures by construction something interesting about  the set of symbols. The drawback is that symbols are altered and, hence, it may be difficult to interpret which symbol is given its distributed representation. In the example, the distributed representations for \emph{eats} and \emph{some} are exactly the same vector $\matrix{W_{3 \times 9}} \vec{e_{5}} = \matrix{W_{3 \times 9}} \vec{e_{6}}$.

Even for distributed representations in the general form, it is possible to define \emph{concatenative composition} and \emph{functional composition} to represent expressions. Vectors $\matrix{W_{d \times n}}\vec{e_i}$ should be replaced to vectors $\vec{e_i}$ in the definition of the concatenative compositionality and the functional compositionality.  
Equation (\ref{conc}) is translated to:
$$
\matrix{Y_s} = \matrix{W_{d \times n}}\vec{conc(s)} = [\matrix{W_{d \times n}}\vec{e}_{w_1} \ldots \matrix{W_{d \times n}}\vec{e}_{w_k}] 
$$
and Equation (\ref{func_sum}) for additive functional compositionality becomes:
$$
\vec{y_s} = \matrix{W_{d \times n}}\vec{func}_{\Sigma}(s)= \sum_{j=1}^k \matrix{W_{d \times n}}\vec{e}_j
$$
In the running example, the additive functional compositionality of sentence $s_1$ is:
$$
\vec{y_{s_1}} = \matrix{W_{3 \times 9}}\vec{func}_{\Sigma}(s_1) = 
\begin{pmatrix}
8\\ 12\\ 0
\end{pmatrix}
$$
Clearly, in this case, it is extremely difficult to derive back the discrete symbolic sequence $s_1$ that has generated the final distributed representation.
 

Summing up, a distributed representation $y_s$ of an discrete symbolic expression $s$ is obtained by using an encoder that acts in two ways: 
\begin{itemize}
\item transforms symbols $w_i$ in vectors by using an embedding matrix $\matrix{W_{d \times n}}$ and the local distributed representation $\vec{e_i}$ of $w_i$;
\item transposes the concatenative compositionality of the discrete symbolic expression $s$ in a functional compositionality by defining the used composition function 
\end{itemize}
When defining a distributed representation, we need to define two elements:
\begin{itemize}
\item an embedding matrix $\matrix{W}$ that should balance these two different aims: (1) \emph{maximize} interpretability, that is, inversion; (2) \emph{maximize} similarity among different symbols for specific purposes.
\item the functional composition model: additive, holographic reduced representations  \citep{Plate1995}, recursive neural networks \cite{Schuster:1997:BRN:2198065.2205129,hochreiter1997long} or with attention \cite{AttentionIsAllYouNeed,BERT}
\end{itemize}
And, the final questions are: What's inside the distributed representation? What's exactly encoded? How this information is used to take decisions?  Hence, the debated question become how concatenative is the functional compositionality in distributed representations behind neural networks?  Can we retrieve discrete symbols and rebuild sequences?

To answer the above questions, we then describe the two properties \emph{Interpretability} and \emph{concatenative compositionality}  for distributed representations. These two properties want to measure how far are distributed representations from symbolic representations.

\textbf{Interpretability} is the possibility of decoding distributed representations, that is, extracting the embedded symbolic representations. This is an important characteristic but it must be noted that it's not a simple yes-or-no classification. It is more a degree associated to specific representations. 
In fact, even if each component of a vector representation does not have a specific meaning, this does not mean that the representation is not interpretable as a whole, or that symbolic information cannot be recovered from it.
For this reason, we can categorize the degree of interpretability of a representation as follows:
\begin{itemize}
\item[] \emph{human-interpretable} -- each dimension of a representation has a specific meaning;
\item[] \emph{decodable} -- the representation may be obscure, but it can be decoded into an interpretable, symbolic representation.
\end{itemize}

\textbf{Concatenative Compositionality for distributed representations} is the possibility of composing basic distributed representations with strong rules and of decomposing back composed representations with inverse rules. Generally, in NLP, basic distributed representations refer to basic symbols.

The two axes of \emph{Interpretability} and \emph{Concatenative Compositionality for distributed representations} will be used to describe the presented distributed representations as we are interested in understanding whether or not a representation can be used to represent structures or sequences and whether it is possible to extract back the underlying structure or sequence given a distributed representation.
It is clear that a local distributed representation is more interpretable than a distributed representation. Yet, both representations lack in concatenative compositionality  when sequences or structures are collapsed in vectors or tensors that do not depend on the length of represented sequences or structures. For example, the bag-of-word local representation does not take into consideration the order of the symbols in the sequence.

\section{Strategies to obtain distributed representations from symbols}

There is a wide range of techniques to transform symbolic representations in distributed representations. 
When combining natural language processing and machine learning, this is a major issue: transforming symbols, sequences of symbols or symbolic structures in vectors or tensors that can be used in learning machines.  These techniques generally propose a function $\eta$ to transform a \emph{local representation} with a large number of dimensions   in a \emph{distributed representation} with a lower number of dimensions: 
$$
\eta \colon \R^n \to \R^d
$$
This function is often called \emph{encoder}.

We propose to categorize techniques to obtain distributed representations in two broad categories, showing some degree of overlapping:
\begin{itemize}
\item representations derived from dimensionality reduction techniques;
\item learned representations
\end{itemize}

In the rest of the section, we will introduce the different strategies according to the proposed categorization. Moreover, we will emphasize its degree of interpretability for each representation and its related function $\eta$  by answering to two questions:
\begin{itemize}
\item Has a specific dimension in $\R^d$ a clear meaning?
\item Can we decode an encoded symbolic representation? In other words, assuming a decoding function $\delta \colon \R^d \to \R^n$, how far is $v \in \R^n$, which represents a symbolic representation, from  $v' = \delta(\eta(v))$?
\end{itemize}
Instead, composability of the resulting representations will be analyzed in Sec. \ref{sec:composing_representations}.

%

\subsection{Dimensionality reductio with Random Projections}
\label{sec:rp}

\emph{Random projection} (RP) \citep{bingham2001random,Fodor02asurvey} is a technique based on random matrices $W_d \in \R^{d \times n}$. Generally, 
%
%
%
the rows of the matrix  $W_d$ are sampled from a Gaussian distribution with zero mean, and normalized as to have unit length \citep{JLL} or even less complex random vectors \citep{achlioptas2003database}. Random projections 
from Gaussian distributions approximately preserves pairwise distance between points (see the \emph{Johnsonn-Lindenstrauss Lemma} \citep{JLL}), that is, for any vector $x, y \in X$:
$$
(1 - \varepsilon)\ \| \vec{x} - \vec{y} \|^2 \leq \| W \vec{x} - W \vec{y} \|^2 \leq (1 + \varepsilon)\ \| \vec{x} - \vec{y} \|^2
$$
where the approximation factor $\varepsilon$ depends on the dimension of the projection, namely, to assure that the approximation factor is $\varepsilon$, the dimension $k$ must be chosen such that:
$$
k \geq \frac{8 \log (m)}{\varepsilon ^ 2}
$$
Constraints for building the matrix $W$ can be significantly relaxed to less complex random vectors \citep{achlioptas2003database}. Rows of the matrix can be sampled from very simple zero-mean distributions such as:
$$
W_{ij} =
\sqrt{3}\begin{cases}
+1 \ \text{ with probability } \frac{1}{6} \\
-1 \ \text{ with probability } \frac{1}{6} \\
0 \ \ \ \text{ with probability } \frac{2}{3} \\
\end{cases}
$$
without the need to manually ensure unit-length of the rows, and at the same time providing a significant speed up in computation due to the sparsity of the projection.



Unfortunately, vectors $\eta(\vec{v})$ are not \emph{human-interpretable} as, even if their dimensions represent linear combinations of dimensions in the original local distribution, these dimensions have not an interpretation or particular properties. 


On the contrary, vectors $\eta(\vec{v})$ are \emph{decodable}. The decoding function is: 
$$
\delta(\vec{v'}) = W_d^T \vec{v'}
$$
and  $W_d^TW_d \approx I$  when $W_d$ is derived using Gaussian random vectors. Hence, distributed vectors in $\R^d$ can be approximately decoded back in the original symbolic representation with a degree of approximation that depends on the distance between $d$ .

The major advantage of RP with respect to PCA is that the matrix $X$ of all the data points is not needed to derive the matrix $W_d$. Moreover, the matrix $W_d$ can be produced \emph{à-la-carte} starting from the symbols encountered so far in the encoding procedure. In fact, it is sufficient to generate new Gaussian vectors for new symbols when they appear. 


\subsection{Learned representation}

Learned representations differ from the dimensionality reduction techniques by the fact that: (1) encoding/decoding functions may not be linear; (2) learning can optimize functions that are different with respect to the target of PCA; and, (3) solutions are not derived in a closed form but are obtained using optimization techniques such as \emph{stochastic gradient decent}. 


Learned representation can be further classified into: 
\begin{itemize}
\item \emph{task-independent representations} learned with a standalone algorithm (as in \emph{autoencoders} \citep{SocherEtAl2011:PoolRAE,Liou201484}) which is independent from any task, and which learns a representation that only depends from the dataset used;
\item \emph{task-dependent representations} learned as the first step of another algorithm (this is called \emph{end-to-end training}), usually the first layer of a deep neural network. In this case the new representation is driven by the task.
\end{itemize}

\subsubsection{Autoencoder}
\label{sec:autoencoders}
Autoencoders are a task independent technique to learn a distributed representation encoder $\eta \colon \R^n \to \R^d$ by using local representations of a set of examples \citep{SocherEtAl2011:PoolRAE,Liou201484}. 
The distributed representation encoder $\eta$ is half of an autoencoder. 

An autoencoder is a neural network that aims to reproduce an input vector in $\R^n$ as output by passing in a hidden layer(s) that are in $\R^d$. Given  $\eta \colon \R^n \to \R^d$ and  $\delta \colon \R^d \to \R^n$ as the encoder and the decoder, respectively, an autoencoder aims to maximize the following function: 
$$
\mathcal{L}(\mathbf{x}, \mathbf{x}') = \| \mathbf{x} - \mathbf{x}' \|^2
$$
where 
$$
\mathbf{x'} = \delta(\eta(\mathbf{x}))
$$
The encoding and decoding module are two neural networks, which means that they are functions depending on a set of parameters $\theta$ of the form
\begin{eqnarray*}
\eta_{\theta}(x) = s(Wx + b) \\
\delta_{\theta'}(y) = s(W'y + b')
\end{eqnarray*}
where the parameters of the entire model are $\theta, \theta' = \left \{ W, b, W', b'\right \} $ with $W, W'$ matrices, $b, b'$ vectors and $s$ is a function that can be either a non-linearity sigmoid shaped function, or in some cases the identity function. In some variants the matrices $W$ and $W'$ are constrained to $W^T = W'$. This model is different with respect to PCA due to the target loss function and the use of non-linear functions.


Autoencoders have been further improved with \emph{denoising autoencoders} \citep{Vincent:2010:SDA:1756006.1953039,vincent2008extracting,masci2011stacked} that are a variant of autoencoders where the goal is to reconstruct the input from a corrupted version. The intuition is that higher level features should be robust with regard to small noise in the input.
In particular, the input $\mathbf{x}$ gets corrupted via a stochastic function:
$$
\tilde{\mathbf{x}} = g(\mathbf{x})
$$
and then one minimizes again the reconstruction error, but with regard to the \emph{original} (uncorrupted) input:
$$
\mathcal{L}(\mathbf{x}, \mathbf{x}') = \| \mathbf{x} -  \delta(\eta(g(\vec{x}))) \|^2
$$
Usually $g$ can be either:
\begin{itemize}
\item adding gaussian noise: $g(\mathbf{x}) = \mathbf{x} + \varepsilon$, where $\varepsilon \sim \mathcal{N}(0, \sigma \mathbb{I})$;
\item masking noise: where a given a fraction $\nu$ of the components of the input gets set to $0$
\end{itemize} 


For what concerns \emph{intepretability}, as for random projection, distributed representations $\eta(\vec{v})$ obtained with encoders from autoencoders  and denoising autoencoders are not \emph{human-interpretable} but are \emph{decodable} as this is the nature of autoencoders.

Moreover, \emph{composability} is not covered by this formulation of autoencoders.

\subsubsection{Embedding layers}

Embedding layers are generally the first layers of more complex neural networks which are responsible to transform an initial local representation in the first internal distributed representation. The main difference with autoencoders is that these layers are shaped by the entire overall learning process. The learning process is generally task dependent. Hence, these first embedding layers depend on the final task. 


It is argued that each layers learn a higher-level representation of its input. This is particularly visible with convolutional network \citep{krizhevsky2012imagenet} applied to computer vision tasks.  In these suggestive visualizations \citep{zeiler2014visualizing}, the hidden layers are seen to correspond to abstract feature of the image, starting from simple edges (in lower layers) up to faces in the higher ones.

However, these embedding layers produce encoding functions and, thus, distributed representations that are not interpretable when applied to symbols. In fact, these distributed representations are not human-interpretable as dimensions are not clearly related to specific aggregations of symbols. Moreover, these embedding layers do not naturally provide decoders. Hence, this distributed representation is  not decodable.

\section{\emph{Distributional} Representations as another side of the coin}

\emph{Distributional} semantics is an important area of research in natural language processing that aims to describe meaning of words and sentences with vectorial representations (see \citep{DBLP:journals/jair/TurneyP10} for a survey). These representations are called \emph{distributional representations}. 

It is a strange historical accident that two similar sounding names -- \emph{distributed} and \emph{distributional} -- have been given to two concepts that should not be confused for many. Maybe, this has happened because the two concepts are definitely related. We argue that distributional representation are nothing more than a subset of distributed representations, and in fact can be categorized neatly into the divisions presented in the previous section

Distributional semantics is based on a famous slogan -- \emph{``you shall judge a word by the company it keeps''} \citep{Firth:1957} -- and on the \emph{distributional hypothesis} \citep{Harris} -- words  have similar meaning if used in similar contexts, that is, words with the same or similar \emph{distribution}. 
Hence, the name distributional as well as the core hypothesis comes from a linguistic rather than computer science background. 

Distributional vectors represent words by describing information related to the contexts in which they appear.  Put in this way it is apparent that a distributional representation \emph{is} a specific case of a distributed representation, and the different name is only an indicator of the context in which this techniques originated. Representations for sentences are generally obtained combining vectors representing words.

Hence, distributional semantics is a special case of distributed representations with a restriction on what can be used as features in vector spaces: features represent a bit of contextual information. Then, the largest body of research is on what should be used to represent contexts and how it should be taken into account. Once this is decided, large matrices $X$ representing words in context are collected and, then, dimensionality reduction techniques are applied to have treatable and more discriminative vectors. 

In the rest of the section, we present how to build matrices representing words in context, we will shortly recap on how dimensionality reduction techniques have been used in distributional semantics, and, finally, we report on \texttt{word2vec} \citep{mikolov2013efficient}, which is a novel distributional semantic techniques based on deep learning.



\subsection{Building distributional representations for words from a corpus}

The major issue in distributional semantics is how to build distributional representations for words by observing word contexts in a collection of documents. In this section, we will describe these techniques using the example of the corpus in Table \ref{corpus}.

\begin{table}[h!]
\begin{center}
\begin{tabular}{cl}
\hline
$s_1$ & \emph{a cat catches a mouse}\\
$s_2$ & \emph{a dog eats a mouse}\\
$s_3$ & \emph{a dog catches a cat}\\
\hline
\end{tabular}
\caption{A very small corpus}
\label{corpus}
\end{center}
\end{table}

A first and simple distributional semantic representations of words is given by word vs. document matrices as those typical in information retrieval \citep{Salton:1989}. Word context are represented by document indexes. Then, words are similar if these words similarly appear in documents. This is generally referred as \emph{topical similarity} \citep{Landauer1997Solution} as words belonging to the same topic tend to be more similar. An example of this approach is given by the matrix in Eq. \ref{first_distributional_representation}. In fact, this matrix is already a distributional and distributed representation for words which are represented as vectors in rows.

A second strategy to build distributional representations for words is to build word vs. contextual feature matrices. These contextual features represent \emph{proxies} for semantic attributes of modeled words \citep{Baroni:2010:DMG:1945043.1945049}. For example, contexts of the word \emph{dog} will somehow have relation with the fact that a dog has four legs, barks, eats, and so on. In this case, these vectors capture a similarity that is more related to a co-hyponymy, that is, words sharing similar attributes are similar. For example, \emph{dog} is more similar to \emph{cat} than to \emph{car} as \emph{dog} and \emph{cat} share more attributes than \emph{dog} and \emph{car}. This is often referred as \emph{attributional similarity} \citep{1174523}.

A simple example of this second strategy are word-to-word matrices obtained by observing n-word windows of target words. For example, a word-to-word matrix obtained for the corpus in Table \ref{corpus} by considering a 1-word window is the following:   
\renewcommand{\kbldelim}{(}
\renewcommand{\kbrdelim}{)}
\begin{equation}
  X = \kbordermatrix{
            & a   & cat & dog & mouse & catches & eats \\
    a       & 0   &  1  &  2  &   2   &    2    &   2  \\
    cat     & 2   &  0  &  0  &   0   &    1    &   0  \\
    dog     & 2   &  0  &  0  &   0   &    1    &   1  \\
    mouse   & 2   &  0  &  0  &   0   &    0    &   0  \\
    catches & 2   &  1  &  1  &   0   &    0    &   0  \\
    eats    & 1   &  0  &  1  &   0   &    0    &   0  \\
  }
\end{equation}
Hence, the word \emph{cat} is represented by the vector $\vec{cat} =  \begin{pmatrix}2   &  0  &  0  &   0   &    1    &   0\end{pmatrix}$ and the similarity between \emph{cat} and \emph{dog} is higher than the similarity between \emph{cat} and \emph{mouse} as the cosine similarity $\cos{\vec{cat}}{\vec{dog}}$ is higher than the cosine similarity  $\cos{\vec{cat}}{\vec{mouse}}$.

The research on distributional semantics focuses on two aspects: (1) the best features to represent contexts; (2) the best correlation measure among target words and features.

How to represent contexts is a crucial problem in distributional semantics. This problem is strictly correlated to the classical question of feature definition and feature selection in machine learning. A wide variety of features have been tried. Contexts have been represented as set of relevant words, sets of relevant syntactic triples involving target words \citep{pado07:_depen,Rothenhausler:2009:UCD:1705415.1705418} and sets of labeled lexical triples \citep{Baroni:2010:DMG:1945043.1945049}.

Finding the best correlation measure among target words and their contextual features is the other issue. Many correlation measures have been tried. The classical measures are \emph{term frequency-inverse document frequency} (\emph{tf-idf}) \citep{Salton:1989} and \emph{point-wise mutual information} ($pmi$). These, among other measures, are used to better capture the importance of contextual features for representing distributional semantic of words.

This first formulation of distributional semantics is a distributed representation that is \emph{interpretable}. In fact, features represent contextual information which is a proxy for semantic attributes of target words \citep{Baroni:2010:DMG:1945043.1945049}.

\subsection{Compacting distributional representations}

As distributed representations, \emph{distributional representations} can undergo the process of dimensionality reduction with Principal Component Analysis and Random Indexing. This process is used for two issues. The first is the classical problem of reducing the dimensions of the representation to obtain more compact representations. The second instead want to help the representation to focus on more discriminative dimensions. This latter issue focuses on the feature selection and merging which is an important task in making these representations more effective on the final task of similarity detection.

Principal Component Analysis (PCA) is largely applied in compacting distributional representations: Latent Semantic Analysis (LSA) is a prominent example \citep{Landauer1997Solution}. LSA were born in Information Retrieval with the idea of reducing word-to-document matrices. Hence, in this compact representation, word context are documents and distributional vectors of words report on the documents where words appear. This or similar matrix reduction techniques have been then applied to word-to-word matrices.

Principal Component Analysis (PCA) \citep{eps273101,pearson1901principal} is a linear method which reduces the number of dimensions by projecting $\R^n$ into the \emph{``best''} linear subspace of a given dimension $d$ by using the a set of data points. The  \emph{``best''} linear subspace is a subspace where dimensions maximize the variance of the data points in the set.  PCA can be interpreted either  as a probabilistic method or as a matrix approximation and is then usually known as \emph{truncated singular value decomposition}. 
We are here interested in describing PCA as probabilistic method as it related to the \emph{interpretability} of the related \emph{distributed representation}.

As a probabilistic method, PCA finds an orthogonal projection matrix $W_d \in \R^{n\times d}$ such that the variance of the projected set of data points is maximized. The set of data points is referred as a matrix  $X\in\R^{m\times n}$  where each row $\vec{x}_i^T \in \R^n$ is a single observation. 
Hence, the variance that is maximized is  $\widehat{X}_d = X W_d^T \in \R^{m \times d}$. 

More specifically, let's consider the first weight vector $\vec{w_1}$, which maps an element of the dataset $\vec{x}$  into a single number $\langle \vec{x}, \vec{w_1} \rangle$. Maximizing the variance means that $\vec{w}$ is such that:
$$
\vec{w_1} = \argmax_{\| \vec{w}\| = 1} \sum_i \left( \langle \vec{x_i}, \vec{w} \rangle \right) ^2
$$
and it can be shown that the optimal value is achieved when $\vec{w}$ is the eigenvector of $X^TX $ with largest eigenvalue.
This then produces a projected dataset:
$$
\widehat{X}_1 =  X^TW_1 = X^T \vec{w_1}
$$
The algorithm can then compute iteratively the second and further components by first subtracting the components already computed from $X$:
$$
X - X\vec{w_1}\vec{w_1}^T
$$
and then proceed as before.
However, it turns out that all subsequent components are related to the eigenvectors of the matrix $X^TX$, that is, the $d$-th weight vector is the eigenvector of $X^TX$ with the $d$-th largest corresponding eigenvalue.

The encoding matrix for distributed representations derived with a PCA method is the matrix:
$$
W_d = \left[ \begin{array}{c}\vec{w}_1\\\vec{w}_2\\ \ldots\\\vec{w}_d\end{array} \right] \in \R^{d \times n}
$$
where $\vec{w}_i$ are eigenvectors with eigenvalues decreasing with $i$.
Hence, local representations $\vec{v} \in \R^n$ are represented in distributed representations in $\R^d$ as:
$$
\eta(\vec{v}) = W_d \vec{v} 
$$

Hence, vectors $\eta(\vec{v})$ are \emph{human-interpretable} as their dimensions represent linear combinations of dimensions in the original local representation and these dimensions are ordered according to their importance in the dataset, that is, their variance. Moreover, each dimension is a linear combination of the original symbols. Then, the matrix  $W_d$ reports on which combination of the original symbols is more important to distinguish data points in the set.

Moreover, vectors $\eta(\vec{v})$ are \emph{decodable}. The decoding function is: 
$$
\delta(\vec{v'}) = W_d^T \vec{v'}
$$
and  $W_d^TW_d = I$ if $d$ is the rank of the matrix $X$, otherwise it is a degraded approximation (for more details refer to \citep{Fodor02asurvey,sorzano2014survey}). Hence, distributed vectors in $\R^d$ can be decoded back in the original symbolic representation with a degree of approximation that depends on the distance between $d$ and the rank of the matrix $X$.

The compelling limit of PCA is that all the data points have to be used in order to obtain the encoding/decoding matrices. This is not feasible in two cases. First, when the model has to deal with big data. Second, when the set of symbols to be encoded in extremely large. In this latter case, local representations cannot be used to produce matrices $X$ for applying PCA.  


In Distributional Semantics, \emph{random indexing} has been used to solve some issues that arise naturally with PCA when working with large vocabularies and large corpora. PCA has some scalability problems:
\begin{itemize}
\item The original co-occurrence matrix is very costly to obtain and store, moreover, it is only needed to be later transformed;
\item Dimensionality reduction is also very costly, moreover, with the dimensions at hand it can only be done with iterative methods;
\item The entire method is not incremental, if we want to add new words to our corpus we have to recompute the entire co-occurrence matrix and then re-perform the PCA step.
\end{itemize}
Random Indexing \citep{sahlgren05} solves these problems: it is an incremental method (new words can be easily added any time at low computational cost) which creates word vector of reduced dimension without the need to create the full dimensional matrix.

Interpretability of compacted distributional semantic vectors is comparable to the interpretability of distributed representations obtained with the same techniques.

\subsection{Learning representations: word2vec}
\label{sec:word2vec}

\begin{figure}
\begin{center}
\includegraphics[width=10cm]{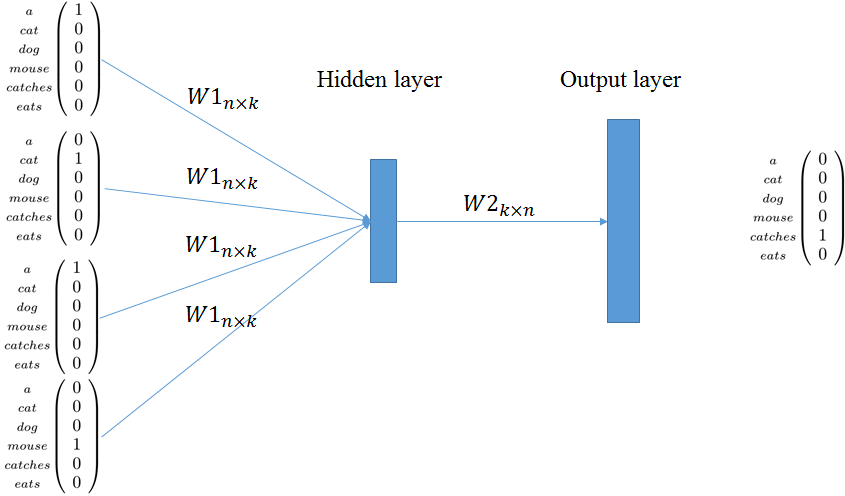}
\end{center}
\caption{word2vec: CBOW model}
\label{CBOW}
\end{figure}

Recently, \emph{distributional hypothesis} has invaded neural networks: \emph{word2vec} \citep{mikolov2013efficient} uses contextual information to learn word vectors. Hence, we discuss this technique in the section devoted to \emph{distributional semantics}.


The name word2Vec comprises two similar techniques, called \emph{skip grams} and \emph{continuous bag of words} (CBOW). Both methods are neural networks, the former takes input a word and try to predict its context, while the latter does the reverse process, predicting a word from the words surrounding it. With this technique there is no explicitly computed co-occurrence matrix, and neither there is an explicit association feature between pairs of words, instead, the regularities and distribution of the words are learned implicitly by the network.

We describe only CBOW because it is conceptually simpler and because the core ideas are the same in both cases. The full  network is generally realized with two layers $W1_{n \times k}$ and $W2_{k \times n}$ plus a softmax layer to reconstruct the final vector representing the word.  In the learning phase, the input and the output of the network are local representation for words. In CBOW, the network aims to predict a target word given context words. For example, given the sentence $s_1$ of the corpus in Table \ref{corpus}, the network has to predict \emph{catches} given its context (see Figure \ref{CBOW}).  

Hence, CBOW offers an encoder $W1_{n \times k}$, that is, a linear word encoder from data where $n$ is the size of the vocabulary and $k$ is the size of the distributional vector. This encoder models contextual information learned by maximizing the prediction capability of the network. A nice description on how this approach is related to previous techniques is given in  \citep{goldberg2014word2vec}.

Clearly, CBOW distributional vectors are not easily human and machine \emph{interpretable}. In fact, specific dimensions of vectors have not a particular meaning and, differently from what happens for auto-encoders (see Sec. \ref{sec:autoencoders}), these networks are not trained to be invertible.

\section{Composing distributed representations}
\label{sec:composing_representations}

In the previous sections, we described how one symbol or a bag-of-symbols can be transformed in distributed representations focusing on whether these distributed representations are \emph{interpretable}. In this section, we want to investigate a second and important aspect of these representations, that is, have these representations \emph{Concatenative Compositionality} as symbolic representations? And, if these representations are \emph{composed}, are still \emph{interpretable}?

\emph{Concatenative Compositionality} is the ability of a symbolic representation to describe sequences or structures by composing symbols with specific rules. In this process, symbols remain distinct and composing rules are clear. Hence, final sequences and structures can be used for subsequent steps as knowledge repositories. 

\emph{Concatenative Compositionality} is an important aspect for any representation and, then, for a distributed representation. Understanding to what extent a distributed representation has \emph{concatenative compositionality } and how information can be recovered is then a critical issue. In fact, this issue has been strongly posed by Plate \citep{Plate1995,Plate:1994} who analyzed how same specific distributed representations encode structural information and how this structural information can be recovered back.

Current approaches for treating distributed/distributional representation of sequences and structures mix two aspects in one model: a \emph{``semantic''} aspect and a \emph{representational} aspect. Generally, the semantic aspect is the predominant and the representational aspect is left aside. For  \emph{``semantic''}  aspect, we refer to the reason why distributed symbols are composed: a final task in neural network applications or the need to give a \emph{distributional semantic vector} for sequences of words. This latter is the case for \emph{compositional distributional semantics} \citep{ClarkCoeckeSadrzadeh2008,baroni2014frege}.  For the \emph{representational} aspect, we refer to the fact that composed distributed representations are in fact representing structures and these representations can be decoded back in order to extract what is in these structures. 

Although the \emph{``semantic''} aspect seems to be predominant in \emph{models-that-compose}, the \emph{convolution conjecture}  \citep{Zanzotto:2015:WGC:2812180.2812190} hypothesizes that the two aspects coexist and the \emph{representational} aspect plays always a crucial role. According to this conjecture, structural information is preserved in any model that composes and structural information emerges back when comparing two distributed representations with dot product to determine their similarity. 

Hence, given the \emph{convolution conjecture}, \emph{models-that-compose} produce distributed representations for structures that can be interpreted back. \emph{Interpretability} is a very important feature in these \emph{models-that-compose} which will drive our analysis.




In this section we will explore the issues faced with the compositionality of representations, and the main ``trends'', which correspond somewhat to the categories already presented. In particular we will start from the work on compositional distributional semantics, then we revise the work on holographic reduced representations \citep{Plate1995,neumann2001holistic} and, finally, we analyze the recent approaches with recurrent and recursive neural networks.
Again, these categories are not entirely disjoint, and methods presented in one class can be often interpreted to belonging into another class.


\subsection{Compositional Distributional Semantics}

In distributional semantics, \emph{models-that-compose} have the name of \emph{compositional distributional semantics models} (CDSMs) \citep{baroni2014frege,Mitchell:Lapata:2010} and aim to apply the principle of compositionality \citep{Frege,Montague:1974} to compute distributional semantic vectors for phrases. These CDSMs produce distributional semantic vectors of phrases by composing distributional vectors of words in these phrases. 
These models generally exploit \emph{structured or syntactic representations} of phrases to derive their distributional meaning.  Hence, CDSMs aim to give a complete semantic model for distributional semantics. 

As in distributional semantics for words, the aim of CDSMs is to produce similar vectors for semantically similar sentences regardless their lengths or structures. For example, words and word definitions in dictionaries should have similar vectors as discussed in \citep{fabio2010-2}.
As usual in distributional semantics, similarity is captured with dot products (or similar metrics) among distributional vectors. 

The applications of these CDSMs encompass multi-document summarization, recognizing textual entailment \citep{DBLP:series/synthesis/2013Dagan} and, obviously, semantic textual similarity detection \citep{agirre-EtAl:2013:*SEM1}.

Apparently, these CDSMs are far from having \emph{concatenative compositionality }, since these distributed representations that can be \emph{interpreted} back. In some sense, their nature wants that resulting vectors forget how these are obtained and focus on the final distributional meaning of phrases. There is some evidence that this is not exactly the case.

The \emph{convolution conjecture} \citep{Zanzotto:2015:WGC:2812180.2812190} suggests that many CDSMs produce distributional vectors where structural information and vectors for individual words can be still \emph{interpreted}. Hence, many CDSMs have the \emph{concatenative compositionality} property and \emph{interpretable}. 

In the rest of this section, we will show some classes of these CDSMs and we focus on describing how these morels are interpretable.



\subsubsection{Additive Models}
\label{sec:additive_models}

\emph{Additive models} for compositional distributional semantics are important examples of \emph{models-that-composes} where \emph{semantic} and \emph{representational} aspects is clearly separated. Hence, these models can be highly \emph{interpretable}.

These additive models have been formally captured in the general framework for two words sequences proposed by  Mitchell\&Lapata \citep{mitchell-lapata:2008:ACLMain}. The general framework for composing distributional vectors of two word sequences  \emph{``u v''} is the following:
\begin{equation}
\mathbf{p}=f(\mathbf{u},\mathbf{v};R;K)\label{eq:(1)-1}
\end{equation}
where $\vec{p} \in \R^n$ is the composition vector, $\vec{u}$ and
$\vec{v}$ are the vectors for the two words \emph{u} and \emph{v}, $R$ is the grammatical relation linking the two words and $K$ is any other additional knowledge used in the composition operation. 
In the additive model, this equation has the following form:
\begin{equation}
\mathbf{p}=f(\mathbf{u},\mathbf{v};R;K) =A_R\mathbf{u}+B_R\mathbf{v}\label{eq:2}
\end{equation}
where $A_R$ and $B_R$ are two square matrices depending on the grammatical relation $R$ which may be learned from data \citep{fabio2010-2,guevara:2010:GEMS}.

\begin{figure}
\begin{center}
\begin{tabular}{cc}
\includegraphics[height=2.5cm]{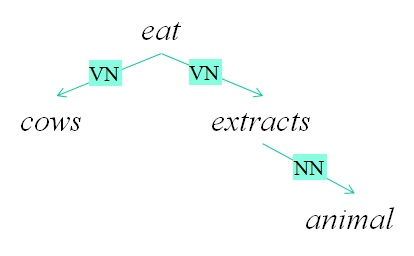}&
\end{tabular}
\end{center}
\caption{A sentence and its dependency graph}
\label{fig:dep_trees}
\end{figure}

Before investigating if these models are interpretable, let introduce a recursive formulation of additive models which can be applied to structural representations of sentences. For this purpose, we use dependency trees. A dependency tree can be defined as a tree whose nodes are words and the typed links are the relations between two words. The root of the tree represents the word that governs the meaning of the sentence. 
A dependency tree $T$ is then a word if it is a final node or it has a root $r_T$ and links $(r_T,R,C_i)$ where $C_i$ is the i-th subtree of the node $r_T$ and $R$ is the relation that links the node $r_T$ with $C_i$. The dependency trees of two example sentences are reported in Figure \ref{fig:dep_trees}. The recursive formulation is then the following:
\begin{displaymath}
f_{r}(T) = \sum_{i} (A_{R} \vec{r_T} + B_{R} f_{r}(C_i)) 
\end{displaymath}

According to the recursive definition of the additive model, the function $f_{r}(T)$ results in a linear combination of elements $M_s \vec{w}_s$ where $M_s$ is a product of matrices that \emph{represents the structure} and $\vec{w}_s$ is the \emph{distributional meaning} of one word in this structure, that is:
\begin{displaymath}
f_{r}(T) = \sum_{s \in S(T)} \matrix{M}_s \vec{w}_s
\end{displaymath}
where $S(T)$ are the relevant substructures of $T$. In this case, $S(T)$ contains the link chains.
For example, the first  sentence in Fig. \ref{fig:dep_trees} has a distributed vector defined in this way:
\begin{eqnarray*}
&&f_{r}(\text{cows eat animal extracts}) =\\
&&= A_{VN}\vec{eat} + B_{VN}\vec{cows} + A_{VN}\vec{eat} + \\
&&+ B_{VN}f_{r}(\text{animal extracts}) =\\
&& = A_{VN}\vec{eat} + B_{VN}\vec{cows} + A_{VN}\vec{eat} +\\
&& + B_{VN}A_{NN}\vec{extracts} + B_{VN}B_{NN}\vec{animal} 
\end{eqnarray*}
Each term of the sum has a part that represents the structure and a part that represents the meaning, for example:
\begin{displaymath}
\overbrace{B_{VN}B_{NN}}^{structure}\underbrace{\vec{beef}}_{meaning}
\end{displaymath}

Hence, this recursive additive model for compositional semantics is a \emph{model-that-composes} which, in principle, can be highly \emph{interpretable}. By selecting matrices $\matrix{M}_s$ such that:
\begin{equation}
\matrix{M}_{s_1}^T\matrix{M}_{s_2} \approx 
\begin{cases}
\matrix{I} & s_1 = s_2 \\
\matrix{0} & s_1 \neq s_2 \\
\end{cases}
\label{eq:JLL_matrices}
\end{equation}
it is possible to recover distributional semantic vectors related to words that are in specific parts of the structure.   For example, the main verb of the sample sentence in Fig. \ref{fig:dep_trees} with a matrix $A_{VN}^T$, that is:
$$
A_{VN}^Tf_{r}(\text{cows eat animal extracts}) \approx 2 \vec{eat}
$$
In general, matrices derived for compositional distributional semantic models \citep{guevara:2010:GEMS,fabio2010-2} do not have this property but it is possible to obtain matrices with this property by applying thee Jonson-Linderstrauss Tranform \citep{JLL} or similar techniques as discussed also in \citep{Zanzotto:2015:WGC:2812180.2812190}.

\subsubsection{Lexical Functional Compositional Distributional Semantic Models}

Lexical Functional Models are compositional distributional semantic models where words are tensors and each type of word is represented by tensors of different order. Composing meaning is then composing these tensors to obtain vectors. These models have solid mathematical background linking Lambek pregroup theory, formal semantics and distributional semantics \citep{DBLP:journals/corr/abs-1003-4394}. Lexical Function models are concatenative compositional, yet, in the following, we will examine whether these models produce vectors that my be \emph{interpreted}.

To determine whether these models produce \emph{interpretable} vectors, we start from a simple Lexical Function model applied to two word sequences. This model has been largely analyzed in \citep{baroni-zamparelli:2010:EMNLP} as matrices were considered better linear models to encode \emph{adjectives}. 

In Lexical Functional models over two words sequences, there is one of the two words which as a tensor of order 2 (that is, a matrix) and one word that is represented by a vector. For example, \emph{adjectives} are matrices and nouns are vectors \citep{baroni-zamparelli:2010:EMNLP} in adjective-noun sequences. Hence, adjective-noun sequences like \emph{``black cat''} or \emph{``white dog''} are represented as:
$$
f(\text{black cat}) = \matrix{BLACK}\vec{cat}
$$
$$
f(\text{white dog}) =\matrix{WHITE}\vec{dog}
$$
where $\matrix{BLACK}$ and $\matrix{WHITE}$ are matrices representing the two adjectives and $\vec{cat}$ and $\vec{dog}$ are the two vectors representing the two nouns.

These two words models are \emph{partially interpretable}: knowing the adjective it is possible to extract the noun but not vice-versa. In fact, if matrices for adjectives are invertible, there is the possibility of extracting which nouns has been related to particular adjectives. For example, if $\matrix{BLACK}$ is invertible, the inverse matrix $\matrix{BLACK}^{-1}$ can be used to extract the vector of \emph{cat} from the vector $f(\text{black cat})$:
$$
\vec{cat} = \matrix{BLACK}^{-1} f(\text{black cat})
$$
This contributes to the \emph{interpretability} of this model. Moreover, if matrices for adjectives are built using Jonson-Lindestrauss Transforms \citep{JLL}, that is matrices with the property in Eq. \ref{eq:JLL_matrices}, it is possible to pack different pieces of sentences in a single vector and, then, select only relevant information, for example:
$$
\vec{cat} \approx \matrix{BLACK}^{T} (f(\text{black cat}) + f(\text{white dog}) )
$$
On the contrary, knowing noun vectors, it is not possible to extract back adjective matrices. This is a strong limitation in term of interpretability.

Lexical Functional models for larger structures are concatenative compositional  but not interpretable at all. In fact, in general these models have tensors in the middle and these tensors are the only parts that can be inverted. Hence, in general these models are not interpretable. However, using the \emph{convolution conjecture} \citep{Zanzotto:2015:WGC:2812180.2812190},  it is possible to know whether subparts are contained in some final vectors obtained with these models.

\subsection{Holographic Representations}

Holographic reduced representations (HRRs) are \emph{models-that-compose} expressly designed to be \emph{interpretable} \citep{Plate1995,neumann2001holistic}. In fact, these models  to encode flat structures representing assertions and these assertions should be then searched in oder to recover pieces of knowledge that is in. For example, these representations have been used to encode logical propositions such as $eat(John,apple)$. In this case, each atomic element has an associated vector and the vector for the compound is obtained by combining these vectors. The major concern here is to build encoding functions that can be decoded, that is, it should be possible to retrieve composing elements from final distributed vectors such as the vector of $eat(John,apple)$.

In HRRs, \emph{nearly orthogonal unit vectors} \citep{JLL} for basic symbols, \emph{circular convolution} $\cc$ and \emph{circular correlation} $\ccorr$ guarantees \emph{composability} and \emph{intepretability}. HRRs are the extension of Random Indexing (see Sec. \ref{sec:rp}) to structures. Hence, symbols are represented with vectors sampled from a multivariate normal distribution $N(0,\frac{1}{d}I_d)$. The composition function is the circular convolution indicated as $\cc$ and defined as:
$$z_j = (\vec{a} \cc \vec{b})_j = \sum_{k=0}^{d-1}a_k b_{j-k}$$
where subscripts are modulo $d$. Circular convolution is commutative and bilinear. This operation can be also computed using \emph{circulant matrices}:
$$\vec{z} = (\vec{a} \cc \vec{b})  = \matrix{A}_{\circ}\vec{b} = \matrix{B}_{\circ}\vec{a}$$
where $\matrix{A}_{\circ}$ and $\matrix{B}_{\circ}$ are circulant matrices of the vectors $\vec{a}$ and $\vec{b}$. Given the properties of vectors $\vec{a}$ and $\vec{b}$, matrices  $\matrix{A}_{\circ}$ and $\matrix{B}_{\circ}$ have the property in Eq. \ref{eq:JLL_matrices}. Hence, \emph{circular convolution} is approximately invertible with the \emph{circular correlation} function ($\ccorr$) defined as follows:
$$c_j = (\vec{z} \ccorr \vec{b})_j = \sum_{k=0}^{d-1}z_k b_{j+k}$$
where again subscripts are modulo $d$. Circular correlation is related to inverse matrices of circulant matrices, that is $\matrix{B}_{\circ}^T$. In the decoding with $\ccorr$, parts of the structures can be derived in an approximated way, that is:
$$(\vec{a} \cc \vec{b}) \ccorr \vec{b} \approx \vec{a}$$

Hence, circular convolution $\cc$ and circular correlation $\ccorr$ allow to build interpretable representations. For example, having the vectors $\vec{e}$, $\vec{J}$, and $\vec{a}$ for $eat$, $John$ and $apple$, respectively, the following encoding and decoding produces a vector that approximates the original vector for $John$:
$$\vec{J} \approx (\vec{J} \cc \vec{e} \cc \vec{a}) \ccorr ( \vec{e} \cc \vec{a})$$ The ``invertibility'' of these representations is important because it allow us not to consider these representations as black boxes.

However, holographic representations have severe limitations as these can encode and decode simple, flat structures. In fact, these representations are based on the circular convolution, which is a commutative function; this implies that the representation cannot keep track of composition of objects where the order matters and this phenomenon is particularly important when encoding nested structures.

Distributed trees \citep{fabio2012-1} have shown that the principles
expressed in holographic representation can be applied to encode
larger structures, overcoming the problem of reliably encoding the order in which elements are composed using the \emph{shuffled circular convolution} function as the composition operator. 
Distributed trees are encoding functions that
transform trees into low-dimensional vectors that also contain the encoding
of every substructures of the tree. Thus, these distributed trees are
particularly attractive as they can be used to represent structures in
linear learning machines which are computationally efficient.

Distributed trees and, in particular, distributed smoothed trees \citep{ferrone-zanzotto:2014:Coling} represent an interesting middle way between compositional distributional semantic models and holographic representation. 

\subsection{Compositional Models in Neural Networks}

When neural networks are applied to sequences or structured data, these networks are in fact \emph{models-that-compose}. However, these models result in \emph{models-that-compose} which are not interpretable. In fact, composition functions are trained on specific tasks and not on the possibility of reconstructing the structured input, unless in some rare cases \citep{SocherEtAl2011:PoolRAE}. The input of these networks are sequences or structured data where basic symbols are embedded in \emph{local} representations or \emph{distributed} representations obtained with word embedding (see Sec. \ref{sec:word2vec}). The output are distributed vectors derived for specific tasks. Hence, these \emph{models-that-compose} are not interpretable in our sense for their final aim and for the fact that \emph{non linear} functions are adopted in the specification of the neural networks. 

In this section, we revise some prominent neural network architectures that can be interpreted as \emph{models-that-compose}: the \emph{recurrent neural networks} \citep{krizhevsky2012imagenet,he2016identity,NIPS2015_5635,DBLP:journals/corr/Graves13} and the \emph{recursive neural networks} \citep{SocherEtAl2012:MVRNN}.

\subsubsection{Recurrent Neural Networks}
Recurrent neural networks form a very broad family of neural networks architectures that deal with the representation (and processing) of complex objects.
At its core a recurrent neural network (RNN) is a network which takes in input the current element in the sequence and processes it based on an internal state which depends on previous inputs.
At the moment the most powerful network architectures are convolutional neural networks \citep{krizhevsky2012imagenet,he2016identity} for vision related tasks and LSTM-type network for language related task \citep{NIPS2015_5635,DBLP:journals/corr/Graves13}.


A recurrent neural network takes as input a sequence $\mathbf{x} = \left( \mathbf{x_1} \ \ldots\ \mathbf{x_n} \right)$ and produce as output a single vector $\mathbf{y} \in \mathbb{R}^n$ which is a representation of the entire sequence.
At each step \footnote{we can usually think of this as a timestep, but not all applications of recurrent neural network have a temporal interpretation} $t$ the network takes as input the current element $\mathbf{x_t}$, the previous output $\mathbf{h_{t-1}}$ and performs the following operation to produce the current output $\mathbf{h_t}$
\begin{eqnarray}
h_t & = &\sigma(W \left[ \mathbf{h_{t-1}}\  \mathbf{x_t}\right] + b)
\end{eqnarray}
where $\sigma$ is a non-linear function such as the logistic function or the hyperbolic tangent and $\left[ \mathbf{h_{t-1}}\  \mathbf{x_t}\right]$ denotes the concatenation of the vectors $  \mathbf{h_{t-1}}$ and $ \mathbf{x_t}$.
The parameters of the model are the matrix $W$ and the bias vector $b$.

Hence, a recurrent neural network is effectively a learned composition function, which dynamically depends on its current input, all of its previous inputs and also on the dataset on which is trained.
However, this learned composition function is basically impossible to analyze or interpret in any way. Sometime an ``intuitive'' explanation is given about what the learned weights represent: with some weights representing information that must be remembered or forgotten.

Even more complex recurrent neural networks as long-short term memory (LSTM) \citep{hochreiter1997long} have the same problem of interpretability. LSTM are a recent and successful way for neural network to deal with longer sequences of inputs, overcoming some difficulty that RNN face in the training phase.
As with RNN, LSTM network takes as input a sequence $\mathbf{x} = \left( \mathbf{x_1} \ \ldots\ \mathbf{x_n} \right)$ and produce as output a single vector $\mathbf{y} \in \mathbb{R}^n$ which is a representation of the entire sequence.
At each step $t$ the network takes as input the current element $\mathbf{x_t}$, the previous output $\mathbf{h_{t-1}}$ and performs the following operation to produce the current output $\mathbf{h_t}$ and update the internal state $\mathbf{c_t}$.
\begin{align*}
f_t  &= \sigma(W_f \left[ \mathbf{h_{t-1}}\  \mathbf{x_t}\right] + b_f)				\\
i_t  &= \sigma(W_i \left[ \mathbf{h_{t-1}}\  \mathbf{x_t}\right] + b_i) 				\\
o_t &= \sigma(W_o \left[ \mathbf{h_{t-1}}\  \mathbf{x_t}\right] + b_o) 			\\
\mathbf{\tilde{c_t}}  &= \tanh(W_c \left[ \mathbf{h_{t-1}}\  \mathbf{x_t}\right] + b_c) 	\\
\mathbf{c_t}  &= f_t \odot \mathbf{c_{t-i}} + i_t \odot \mathbf{\tilde{c_t}} 			\\
h_t  &= o_t \odot \tanh(\mathbf{c_t}) 									
\end{align*}
where $\odot$ stands for element-wise multiplication, and the parameters of the model are the matrices $W_f, W_i, W_o, W_c$ and the bias vectors $b_f, b_i, b_o, b_c$.

Generally, the interpretation offered for recursive neural networks is \emph{functional} or \emph{``psychological''} and not on the content of intermediate vectors. For example, an interpretation of the parameters of LSTM is the following:
\begin{itemize}
\item $f_t$ is the \emph{forget gate}: at each step takes in consideration the new input and output computed so far to decide which information in the internal state must be \emph{forgotten} (that is, set to $0$);
\item $i_t$ is the \emph{input gate}: it decides which position in the internal state will be updated, and by how much;
\item $\tilde{c_t}$ is the proposed new internal state, which will then be updated effectively combining the previous gate;
\item $o_t$ is the \emph{output gate}: it decides how to modulate the internal state to produce the output
\end{itemize}


These \emph{models-that-compose} have high performance on final tasks but are definitely not interpretable.

\subsubsection{\label{sub:Recursive-Neural-Network}Recursive Neural Network}

\begin{figure}
\begin{center}
\begin{small}
\tikzset{level distance=18pt}
\Tree[.S [.cows ] [.VP [.eat ] [.NP [.animal ] [.extracts ] ] ] ]
\end{small}
\end{center}
\caption{A simple binary tree}
\label{fig:binary_tree}
\end{figure}

\begin{figure}
\begin{center}
\includegraphics[height=4.5cm]{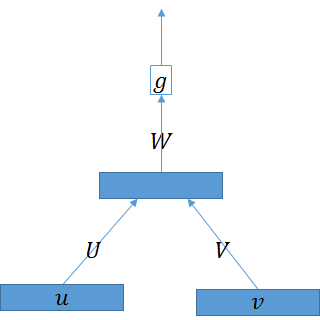}
\end{center}
\caption{Recursive Neural Networks}
\label{fig:recnn}
\end{figure}

The last class of \emph{models-that-compose} that we present is the class of \emph{recursive neural networks} 
\citep{SocherEtAl2012:MVRNN}. These networks are applied to data structures as trees and are in fact applied recursively on the structure. Generally, the aim of the network is a final task as \emph{sentiment analysis} or \emph{paraphrase detection}.

Recursive neural networks is then a basic block (see Fig. \ref{fig:recnn}) which is recursively applied on trees like the one in Fig. \ref{fig:binary_tree}. The formal definition is the following:
\[
\mathbf{p}=f_{U,V}(\mathbf{u},\mathbf{v})=f(V\mathbf{u},U\mathbf{v})=g(W\begin{pmatrix}V\mathbf{u}\\
U\mathbf{v}
\end{pmatrix})
\]
where $g$ is a component-wise sigmoid function or $\mathrm{tanh}$,
and $W$ is  a matrix that maps the concatenation vector$\begin{pmatrix}V\mathbf{u}\\
U\mathbf{v}
\end{pmatrix}$  to have the same dimension. 

This method deals naturally with recursion: given a binary parse tree of a sentence $s$, the algorithm creates vectors and matrices representation for each node, starting from the terminal nodes. Words are represented by distributed representations or local representations.  For example, the tree in Fig. \ref{fig:binary_tree} is processed by the recursive network in the following way. First, the network in Fig. \ref{fig:recnn} is applied to the pair \emph{(animal,extracts)} and $f_{UV}(\vec{animal},\vec{extract})$ is obtained. Then, the network is applied to the result and \emph{eat} and $f_{UV}(\vec{eat},f_{UV}(\vec{animal},\vec{extract}))$ is obtained and so on.

Recursive neural networks are not easily interpretable even if quite similar to the additive \emph{compositional distributional semantic models} as those presented in Sec. \ref{sec:additive_models}. In fact, the non-linear function $g$ is the one that makes final vectors less interpretable.

\section{Conclusions}

Natural language is symbolic representation. Thinking to natural language understanding systems which are not based on symbols seems to be extremely odd. However, recent advances in machine learning (ML) and in natural language processing (NLP) seem to contradict the above intuition: symbols are fading away, erased by vectors or tensors called \emph{distributed} and \emph{distributional representations}.

We made this survey to show the not-surprising link between symbolic representations and distributed/distributional representations. This is the right time to revitalize the area of interpreting how symbols are represented inside neural networks. In our opinion, this survey will help to devise new deep neural networks that can exploit existing and novel symbolic models of classical natural language processing tasks. We believe that a clearer understanding of the strict link between distributed/distributional representations and symbols may lead to radically new deep learning networks.

\bibliographystyle{frontiersinSCNS_ENG_HUMS}
\bibliography{LSTM,refined_bibliography,additional}

\end{document}